\documentclass[11pt]{article}

\usepackage{acl}

\usepackage{times}
\usepackage{latexsym}

\usepackage[T1]{fontenc}

\usepackage[utf8]{inputenc}

\usepackage{microtype}

\usepackage{inconsolata}
\usepackage{url}
\usepackage{amssymb}
\usepackage{graphicx}
\usepackage{booktabs}
\usepackage{CJKutf8}
\usepackage{graphicx}
\usepackage{subfigure}
\usepackage{float}
\usepackage{amsmath}
\usepackage{amssymb}
\usepackage{multirow}
\usepackage{booktabs}
\usepackage{url}
\usepackage{array}
\usepackage{enumitem}
\usepackage{algorithm}
\usepackage{algorithmic}
\usepackage{pifont}
\usepackage{bm}
\usepackage{lipsum}
\usepackage{xcolor}
\usepackage{makecell}
\usepackage{CJKutf8}
\usepackage{tikz}
\usepackage[edges]{forest}
\usepackage{natbib}
\definecolor{hidden-draw}{RGB}{20,68,106}
\definecolor{hidden-pink}{RGB}{255,245,247}

\usepackage[many]{tcolorbox}
\definecolor{main}{HTML}{5989cf}    
\definecolor{sub}{HTML}{cde4ff}     
\newtcolorbox{boxB}{
    enhanced, 
    boxrule = 0pt, 
    borderline = {0.75pt}{0pt}{main}, 
    borderline = {0.75pt}{2pt}{sub} 
}
\newtcolorbox{boxH}{
    colback = sub, 
    colframe = main, 
    boxrule = 0pt, 
    leftrule = 3pt 
}

\title{Data Management For Training Large Language Models: A Survey}

\author{
\textbf{Zige Wang}$^{1,2}$\thanks{ \quad Work done during Zige Wang's internship at Huawei Noah's Ark Lab.} \quad \textbf{Wanjun Zhong}$^{2}$\thanks{ \quad Corresponding author (zhongwanjun1@huawei.com)} \quad \textbf{Yufei Wang}$^{2}$ \quad \textbf{Qi Zhu}$^{2}$ \quad \textbf{Fei Mi}$^{2}$ \quad \textbf{Baojun Wang}$^{2}$ \\
\textbf{Lifeng Shang}$^{2}$ \quad \textbf{Xin Jiang}$^{2}$ \quad \textbf{Qun Liu}$^{2}$\\
$^{1}$School of Computer Science, Peking University \\
$^{2}$Huawei Noah's Ark Lab \\
zigewang@stu.pku.edu.cn \\
\{zhongwanjun1, wangyufei44, zhuqi41, mifei2, puking.w\}@huawei.com \\
\{Shang.Lifeng, Jiang.Xin, qun.liu\}@huawei.com \\
}

\begin{document}
\maketitle
\begin{abstract}
Data plays a fundamental role in training Large Language Models (LLMs). Efficient data management, particularly in formulating a well-suited training dataset, is significant for enhancing model performance and improving training efficiency during pretraining and supervised fine-tuning stages. Despite the considerable importance of data management, the underlying mechanism of current prominent practices are still unknown. Consequently, the exploration of data management has attracted more and more attention among the research community. This survey aims to provide a comprehensive overview of current research in data management within both the pretraining and supervised fine-tuning stages of LLMs, covering various aspects of data management strategy design. Looking into the future, we extrapolate existing challenges and outline promising directions for development in this field. Therefore, this survey serves as a guiding resource for practitioners aspiring to construct powerful LLMs through efficient data management practices.

\end{abstract}

\section{Introduction}
Large Language Models (LLMs) have shocked the natural language processing (NLP) community with their strong performance and emergent abilities~\cite{openai2023gpt4, touvron2023llama, wei2022emergent}. According to previous studies~\cite{kaplan2020scaling, hoffmann2022empirical}, LLMs' achievements depend heavily on self-supervised pretraining over processed vast volumes of text data. Recent research~\cite{zhou2023lima, ouyang2022training} further enhances LLMs' instruction-following ability and performance on downstream tasks through Supervised Fine-Tuning (SFT) on deliberately curated instruction datasets.

To construct suitable training datasets, data management is vitally important and challenging in both the pretraining and SFT stages of LLMs, which we define as following:

\begin{boxB}
    \textbf{Data management:} the process of organizing a well-suited training dataset with collected data, including the data selection, combination and utilization strategies, and the evaluation of the chosen strategies.
\end{boxB}

In the pretraining stage, constructing datasets with high-quality data is essential for efficient training~\cite{jain2020overview, gupta2021data}. To equip LLMs with diverse and comprehensive abilities, heterogeneous dataset composition with mixtures of domains is also required~\cite{gao2020pile, longpre2023pretrainer, shen2023slimpajama}. However, many prominent LLMs do not enclose~\cite{anil2023palm, openai2023gpt4} or only document~\cite{brown2020language, le2023bloom, touvron2023llama} the techniques used in the construction of their pretraining dataset, leaving the reasons and effects of choosing specific data management strategies absent. In the SFT stage, LLMs' performance and instruction-following abilities are primarily evoked by carefully constructed instruction datasets~\cite{sanh2022multitask, ouyang2022training}. Although a handful of instruction datasets/benchmarks have been proposed~\cite{wang2022super, wang-etal-2023-self-instruct, alpaca, gpt4all}, practitioners still find it confusing about the effects of instruction datasets on the performance of fine-tuned LLMs, leading to difficulties in choosing proper data management strategies in LLM SFT practices. To address the sparsity problem of existing data, collecting data from multimodal source~\cite{zhang2023llama, yang2023gpt4tools} and model synthesis~\cite{maini2024rephrasing, li2024synthetic} rise as new trends. 

To address these challenges, researchers try to discover and explore the underlying principles of data management.  
With more and more works been proposed to address different aspects, it is necessary to conduct a systematic discussion considering the whole picture. This survey aims to provide a comprehensive overview of current research in LLM data management and a guiding resource to practitioners attempting to build powerful LLMs with efficient data management practices.

In Section~\ref{sec:pretraining} and~\ref{sec:sft}, we respectively discuss current research in the pretraining and SFT stages of LLMs, covering multiple aspects in data management like domain/task composition, data quality, data quantity, etc., as shown in Figure~\ref{fig:framework}. 
However, there still lacks a well-established and acknowledged general data management pipeline. Hence, We hope our work can inspire future research to establish and analyze such general pipelines. With the vision that the development of data management should keep pace with that of LLMs' abilitites, we present more existing challenges and promising future directions in Section~\ref{sec:challenges}.

\begin{figure*}[t]
    \centering
    \includegraphics[width=\textwidth]{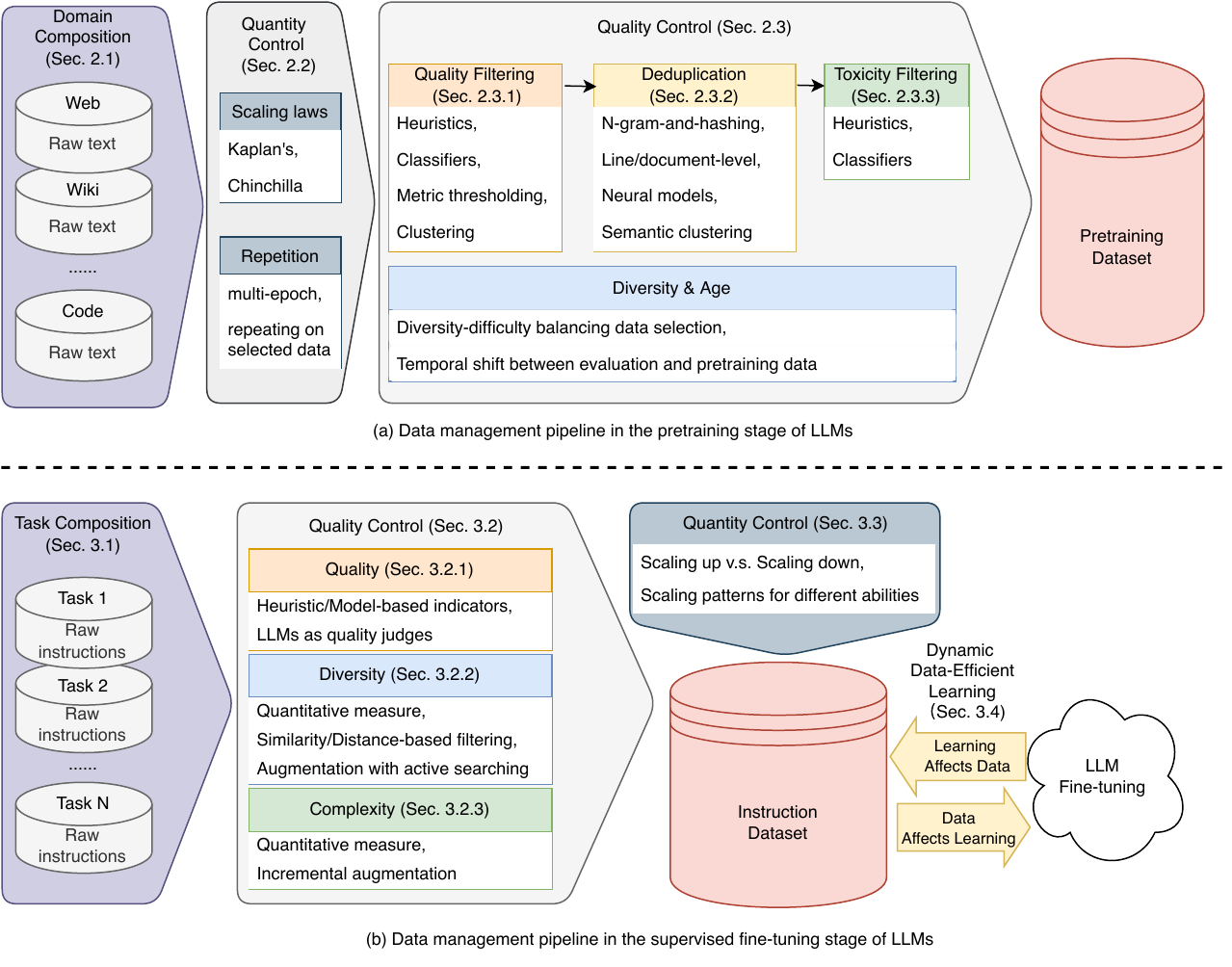}
    \caption{Data management pipelines for the pretraining and supervised fine-tuning of Large Language Models.}
    \label{fig:pipelines}
    \vspace{-5mm}
\end{figure*}

\section{Pretraining of LLM}
\label{sec:pretraining}

Data management is found to be important in the pretraining stage of many prominent LLMs~\cite{openai2023gpt4, touvron2023llama, wei2022emergent}.
 
In this section, we will discuss works trying to explore data management in the pretraining stage of LLMs, including domain composition, data quantity and data quality, as shown in Figure~\ref{fig:pipelines}(a). 
Strategies adopted by prominent pretrained models are listed in Table~\ref{tab:pretrain}.

\subsection{Domain Composition}
\label{subsec:pt-composition}

\begin{figure*}[t]
    \centering
    \includegraphics[width=0.7\textwidth]{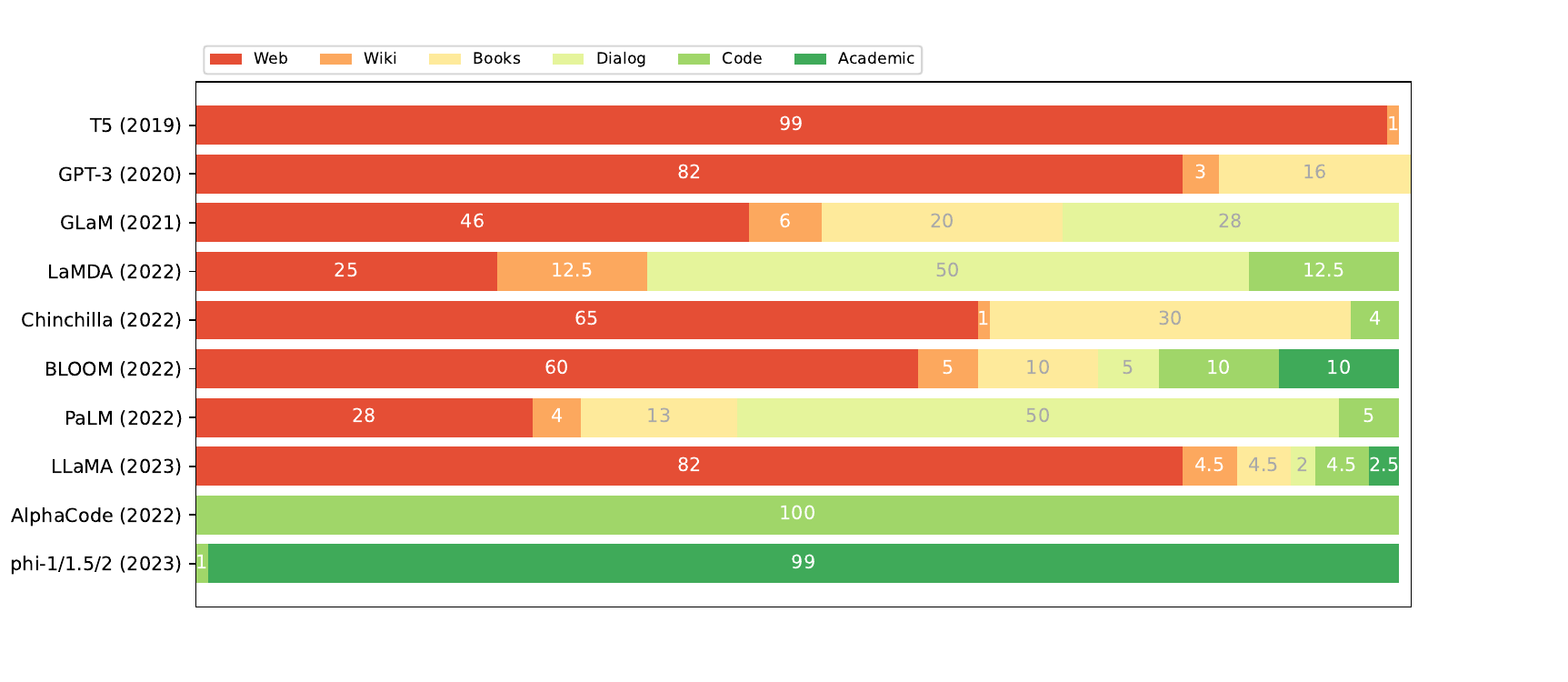}
    \caption{The domain composition of prominent Large Language Models.}
    \label{fig:domains}
    \vspace{-5mm}
\end{figure*}

Publicly available pretraining datasets, like the Pile~\cite{gao2020pile}, usually contain mixtures of data collected from multiple sources and domains. Many prominent models~\cite{du2022glam, gao2023llama, zhang2023llama} are also trained on a mixture of data from different domains. Figure~\ref{fig:domains} summarizes the revealed domain mixture ratios in the pretraining datasets of prominent models. 

Early pretraining corpus mostly contain data with high diversity (Web and Wiki). With recent emphasis on the data quality and the requirement for advanced abilities, high quality text (Books and academic text) are integrated. Most recently, with improved importance of Coding LLMs and the essential finding that code-based pretraining can enhance reasoning capability of LLMs~\cite{liang2022holistic, guo2024deepseek}, domain data like code and math take up higher ratio of the total pretraining data. A trend can be concluded that more and more domains are included to pretrain LLMs with more various and powerful abilities. The benefits of multi-domain composition are also studied in a recent study~\cite{longpre2023pretrainer}.

Proper domain mixture ratio is also important in the pretraining of LLMs. Early attempts usually found the ratio by elaborated experiments and intuitions~\cite{gao2020pile, du2022glam, thoppilan2022lamda}. Recently, domain generalization techniques are leveraged to automatically assign domain weights to form a suitable target distribution, such as importance resampling~\cite{xie2023data} and Group Distributionally Robust Optimization~\cite{xie2023doremi}. Contribution of each domain measured via gradients is also adopted to reweight domains~\cite{fan2023doge}. \citet{xia2023sheared} assign batch-level weights dynamically based on varying losses. \citet{ye2024data} propose data mixing laws to predict model performance with different mixing ratios.

Although proper domain composition is broadly acknowledged as beneficial in the pretraining of LLMs, as discussed previously, some empirical analyses arrive at different conclusions and leave open questions for future research. For example, \citet{longpre2023pretrainer}
claim that the inclusion of diverse web domains may perform better than specific mixtures in certain tasks. \citet{nijkamp2023codegen2} study programming and natural language mixtures and find that models trained with mixtures do not perform better than but closely to domain-matched models given the same computing budget.

\vspace{-2mm}
\subsection{Data Quantity}
\label{subsec:pt-quantity}
It is recognized that the pretraining of LLMs needs large amounts of data. Scaling laws are proposed to depict the relationships between data quantity and model size. Repeatedly training on data is also studied due to data exhaustion. 

\subsubsection{Scaling Laws}
Before the popularization of LLMs, the relationship between training dataset size and the performance of Transformer-based language models~\cite{vaswani2017attention} had already attracted researchers' attention. \citet{kaplan2020scaling} find that the language model loss has a power-law relationship with training dataset size or model size, respectively, when not bottlenecked by each other and the training computing budget. They further depict the dependence between model size and training dataset size as:

\begin{equation} \label{eq-kaplan3}
    L(N, D) = \big [ \big (\frac{N_c}{N} \big )^{\frac{\alpha_N}{\alpha_D}} + \frac{D_c}{D} \big ]^{\alpha_D}
\end{equation}

where $L$ is the language model test loss, $D$ is the number of training tokens, $N$ is the number of model parameters, $\alpha_D$ and $\alpha_N$ are the power-law components for the scaling of $D$ and $N$, respectively, and $D_c$ and $N_c$ are constant numbers~\footnote{The precise numerical values of $D_c$ and $N_c$ depend on vocabulary size and tokenization and do not have fundamental meaning.}.

Fitting Equation~\ref{eq-kaplan3}, they conclude that model loss decreases predictably as long as the model size and training dataset size are scaled up simultaneously. Still, overfitting will happen if either of them is fixed while the other increases.  Given fixed computing budget $C$, they analyze the optimal allocation of $D_{opt} \sim C^{0.27}$ and $N_{opt} \sim C^{0.73}$, showing that the model size should increase faster than the training dataset size.

Following \citet{kaplan2020scaling}, \citet{hoffmann2022empirical} conduct experiments on much larger language models and arrive at a new scaling law, usually called as \textit{Chinchilla Scaling Law}:

\begin{equation}
    L(N, D) = E + \frac{A}{N^\alpha} + \frac{B}{D^\beta}
\end{equation}

where they empirically fit $E=1.69$, $A=406.4$, $B=410.7$, $\alpha = 0.34$ and $\beta = 0.28$. The optimal allocation of $D_{opt}$ and $N_{opt}$ are also analyzed as $D_{opt} \sim C^{0.54}$ and $N_{opt} \sim C^{0.46}$. Hence, they draw a different conclusion that model and training dataset sizes should scale roughly at the same rate with a larger computing budget. \citet{su2024unraveling} dig deeper into Kaplan's scaling laws and provide more detailed instructions to fit the constants.

\subsubsection{Data Repetition}

While \citet{kaplan2020scaling} and \citet{hoffmann2022empirical} both focus on scaling laws with unique data trained only for one epoch, ~\citet{hernandez2022scaling} study the scaling laws with a small fraction of repeated data in the training dataset and find that the text overlap may be harmful to model performance, causing a divergence from Kaplan's scaling law on model size larger than 100M parameters. 

With the models grow larger and larger, data has becoming more and more demanding, raising concerns about the exhaustion of high-quality training data~\cite{villalobos2022will, hoffmann2022empirical}. 
Addressing these concerns, several works study the consequence of repeatedly pretraining on the whole datasets for multiple epochs. Scaling law on repeated training is proposed to depict the diminishing of returns with more repetition and larger model sizes~\cite{muennighoff2023scaling} and shows a multi-epoch degradation phenomenon~\cite{xue2023repeat}. Further analysis digs out that dataset size, model parameters, and training objectives are the key factors to this phenomenon, and classic regularization techniques may not be helpful, except for dropout~\cite{xue2023repeat}. 

There are still positive results in the research of data repetition. \citet{muennighoff2023scaling} find that repeatedly training on the whole dataset up to 4 epochs only causes trivial harm to test loss compared to training on unique new data. Instead of simply repeating over the whole dataset, \citet{tirumala2023d4} show that repeatedly training on carefully selected data can outperform that on randomly selected new data, suggesting a feasible way of repeating on intelligently selected data.

Recently, pretraining with mixed real and synthesized data is adopted to meet the data exhaustion challenge~\cite{2023phi2, meta2024llama3}. It is also gaining more an more attention and develops into a new trend as data synthesize.

\subsection{Data Quality}
\label{subsec:pt-quality}
In the pretraining of LLMs, Quality control techniques of the pretraining datasets usually form an order~\cite{rae2021scaling, nguyen2023culturax, tirumala2023d4, gan2023ziya2}, namely quality filtering, deduplication and toxicity filtering. Data diversity and age are also explored.

\subsubsection{Quality Filtering}
Public datasets like Common Crawl~\footnote{https://commoncrawl.org/, a large text corpus contains raw web page data, metadata extracts, and text extracts.} and multilingual datasets~\cite{kreutzer-etal-2022-quality} usually contain low-quality data that hampers the training of LLMs. Hence, existing works usually perform quality filtering using hand-crafted heuristics~\cite{Yang2019XLNetGA, raffel2020exploring, nijkamp2022codegen}, a trained classifier~\cite{brown2020language, gao2020pile, du2022glam, touvron2023llama, wettig2024qurating}, metric thresholding~\cite{wenzek2020ccnet, muennighoff2023scaling} or combinations of these techniques. Besides instance-level filtering, embedding clustering is also adopted to filter one cluster at a time~\cite{kaddour2023minipile}.

Despite the reduction of data quantity, quality filtering is usually proven to be beneficial in model performance improvement~\cite{longpre2023pretrainer}. Several carefully filtered high-quality datasets are proposed to train lightweight LLMs with outstanding performances~\cite{gunasekar2023textbooks, li2023textbooks, 2023phi2, penedo2023refinedweb}.
However, \citet{gao2021empirical} finds that aggressive filtering might lead to performance degradation on a wide range of tasks for GPT-like LLMs due to the poor representativity of the filtering proxy objectives. To address this issue, \citet{marion2023less} comprehensively examine different data quality estimators and find that pruning datasets based on perplexity performs better than more complicated techniques like memorization. \citet{gan2023ziya2} develop data-centric scaling laws and show that improving semantic and grammatical quality is more effective. However, there still lacks a well-established and theoretically efficient filtering strategy, leaving room for further exploration.

\subsubsection{Deduplication}
Deduplication is a necessary step in many LLMs' pretraining data management procedures and the preprocessing of many publicly available datasets~\cite{brown2020language, le2023bloom, touvron2023llama, raffel2020exploring}. \citet{lee2021deduplicating} find that deduplication is beneficial in memorization mitigation, train-test overlap avoidance, and training efficiency improvement while keeping model perplexity. \citet{kandpal2022deduplicating} also show that deduplication can considerably lower the success rate of privacy attacks aiming at model memorization.

Among practices of deduplication, N-gram-and-hashing is the most commonly adopted technique~\cite{lee2021deduplicating, borgeaud2022improving, rae2021scaling}. It can operate at line-level~\cite{touvron2023llama}, document-level~\cite{hoffmann2022empirical, li2022competition} or combinations of them. Recently, neural models are experimentally proven to outperform traditional N-gram-and-hashing methods~\cite{silcock2022noise}. Addressing semantic deduplication, \citet{abbas2023semdedup} propose \textit{SemDeDup} to remove semantic duplicates that lie closely in the pretrained model's embedding space and apply clustering to reduce the searching computation.

\subsubsection{Toxicity Filtering}
\label{subsubsec:toxic}
Toxicity refers to the text content which is \textit{"rude, disrespectful, or unreasonable language that is likely to make someone leave a discussion"}~\cite{gehman2020realtoxicityprompts, welbl2021challenges}. As raw text corpora usually contain toxic text~\cite{luccioni2021s, longpre2023pretrainer}, toxicity filtering aims to remove text with undesirable toxic text in the pretraining datasets, further preventing LLMs from generating toxic utterances. Similar to quality filtering, heuristic and rule-based filtering~\cite{lees2022new, gargee2022analyzing, friedl2023dis} and N-gram classifiers~\cite{raffel2020exploring} are usually adopted as toxicity filters. 

Although effective in model detoxifying, \citet{longpre2023pretrainer} discover that toxicity filtering reduces the risk of toxic generation by sacrificing model generalization and toxicity identification ability. Moreover, \citet{xu2021detoxifying} and \citet{welbl2021challenges} find that training dataset detoxification leads to the marginalization of minority groups like dialects and minority identity mentions, posing challenges in building unbiased LLMs.

\subsubsection{Data Diversity}
Some works focus on other aspects of data management in the pretraining stage of LLMs. \citet{lee2023beyond} show that the format diversities of publicly available pretraining datasets are high when measured by Task2Vec diversity coefficient~\cite{miranda2022curse}. 
\citet{maharana2023d2} propose \textit{D2 Pruning} to balance data diversity and difficulty in data selection by representing datasets as undirected graphs and adopting forward-and-reverse message passing strategy to select a subgraph enveloping both diverse and difficult data samples.

\subsubsection{Data Age}
In current practices, more recent LLMs are usually pretrained using newer data~\footnote{https://platform.openai.com/docs/models}. Some knowledge learned by pretrained LLMs could also be time-sensitive. 
\citet{longpre2023pretrainer} study the impact of data age and find that the temporal shift between evaluation and pretraining data will lead to inaccurate performance estimation. This temporal misalignment might not be overcome by fine-tuning, especially for larger models.

\subsection{Relations Among Domain Composition, Data Quantity and Data Quality}
\label{sec:pt-relation}
Recently, several scaling laws are proposed to explore the synergistic effect of different aspects on the pretrained model performance, such as the bivariate model performance prediction regarding data quantity and domain composition ratio~\cite{ge2024data}, the quality-quantity tradeoff under different computing budget~\cite{goyal2024scaling}, and the positive correlation between data quality and model scale under the same data quantity~\cite{bi2024deepseek}.
What's more, \citet{shen2023slimpajama} emphasize global deduplication to remove overlaps among different domains. \citet{longpre2023pretrainer} claim that domains with high quality and diversity are more beneficial than other domains.

\section{Supervised Fine-Tuning of LLM}
\label{sec:sft}

Based on the general knowledge and capabilities learned in the pretraining stage, supervised fine-tuning (SFT) is proposed to further improve LLMs with instruction-following ability and alignment with human expectations~\cite{wei2021finetuned, sanh2022multitask, ouyang2022training}. 
Although LLMs fined-tuned with existing instruction datasets have achieved remarkable performance in various NLP tasks, the impacts of instruction data management on fine-tuned models are still under debate. The data management process in the SFT stage can be summarized as illustrated in Figure~\ref{fig:pipelines}(b), including task composition, data quality control, data quantity control and dynamic data-efficient learning. Table~\ref{tab:sft} summarizes the data management practices of prominent fine-tuned LLMs.

\subsection{Task Composition}
\label{subsec:sft-composition}

Since LLMs have shown surprisingly emergent abilities in handling various NLP tasks, multitask fine-tuning appears to be promising to improve LLMs' generalization performance on unseen tasks. The benefits of increasing the number of tasks in SFT have been experimentally proven on models with different sizes ranging from 3B to 540B parameters~\cite{wang2022super, sanh2022multitask, wei2021finetuned, chung2022scaling}. With the scaling of tasks, the mixture ratio of data targeting different tasks is also found to be critical and usually decided by experiments and intuitions~\cite{iyer2022opt, longpre2023flan}. To enable LLMs to solve targeted tasks with specific skills, representation similarity~\cite{ivison-etal-2023-data, lee2024instruction} and gradient similarity~\cite{xia2024less} is proposed to select relevant multitask subsets. 

However, conflicts might exist among the many tasks. \citet{dong2023abilities} focus on task composition among mathematical reasoning, code generation, and general human-aligning abilities. They find that model abilities are improved when the mixed data amount is small but decreased otherwise. The negative impact of large amount mixing data might lie in the similarity degree of data format and data distribution among different SFT tasks. \citet{wang2023far} also experimentally show that different instruction datasets may correspond to different specific abilities. And winning across all evaluations using a single dataset or combination seems to be challenging.

Divergent from compositing multiple tasks, some works claim that integration of LLMs tuned on single task data can outperform one LLM tuned on multiple tasks~\cite{jang2023exploring, chen2023maybe}. But fine-tuning more task-specific LLMs also means more resource consumption. How to efficiently equip LLMs with the ability to solve multiple tasks still demands more exploration.

\subsection{Data Quality}
\label{subsec:sft-quality}
Data quality is always a focal point in the SFT of LLMs, addressing instruction quality, diversity, and complexity. Here, we focus more on managing and analyzing existing instruction data instead of instruction generation methods discussed in previous surveys~\cite{zhang2023instruction, wang2023aligning}.

\subsubsection{Instruction Quality}
Many researchers have found that the quality of instruction data is one of the most important factors in improving model performance~\cite{chia2023instructeval, zhou2023lima, ding2023enhancing}. 
During the construction of instruction dataset, there is usually a filtering step to select high-quality instructions generated by models. 

Heuristic- and model-based natural language indicators like perplexity and uncertainty are commonly adopted filtering criteria~\cite{wang2023harnessing, cao2023instruction, Bhatt2024AnED}. What's more, losses~\cite{zhou2023lobass, Li2023FromQT, Li2024SelectiveRS} and output probabilities~\cite{Li2023TunaIT, Li2023OneSL, chen2024automated, he2024shed, liu2024selectit} of LLMs are adopted to compute more complex scores for data selection. Popular searching approaches like BlendSearch~\cite{wang2020economic} are also leveraged to find high-quality instructions satisfying the criteria~\cite{cao2023instruction}.

In addition, LLMs are also queried to directly evaluate the quality of instructions. Fine-tuned LLMs are prompted to assign quality scores~\cite{li2023self} or provide self-feedback~\cite{lu2023self, madaan2023self} to their own responses to iteratively improve model prediction. Strong LLMs like ChatGPT~\cite{selfee2023, chen2023alpagasus, Li2023TunaIT} or reward models~\cite{Du2023MoDSMD} are also adopted as quality judges during instruction data filtering. Recently, weak-to-strong strategy is introduced to select high-quality data with smaller and weaker models~\cite{Li2024SuperfilteringWD, yang2024smalltolarge, mekala2024smaller}.

\subsubsection{Instruction Diversity}
The intention and semantic diversity of instructions is another important factor that has shown positive effects on model performance improvement and robustness~\cite{zhou2023lima, ding2023enhancing, alpaca, Bukharin2023DataDM}. However, there is no well-acknowledged measurement to quantitatively indicate the diversity of an instruction dataset. \textit{\#InsTag}~\cite{lu2023instag} propose to measure instruction diversity using fine-grained tags generated by ChatGPT~\footnote{https://chatgpt.openai.com/}. Specifically, it quantifies instruction diversity as the unique tag coverage rate in the overall tag set. 

To maintain both diversity and data-efficiency in the instruction datasets, Rouge-L similarity~\cite{wang-etal-2023-self-instruct}, embedding distance~\cite{wu2023self, Bukharin2023DataDM, huang2024multi} and scoring models~\cite{ge2024clustering} are proposed to select instructions that are different from each other in literal, semantic and human-aligning level.

Due to data constraints, diversity can be challenging in some domain-specific tasks. Thus, \citet{wan2023explore} propose to enlarge the data coverage through active searching variations and possibilities of instructions using LLMs.

\subsubsection{Instruction Complexity}
Instruction complexity is found to be crucial in developing LLMs with complex instruction-following and reasoning abilities~\cite{xu2023wizardlm, luo2023wizardcoder, mukherjee2023orca, he2024complex}. Several works endeavor to quantify and evaluate instruction complexity. Using aforementioned tags, \textit{\#InsTag}~\cite{lu2023instag} quantifies complexity as the average tag number assigned to each query in a dataset. \citet{he2023can} evaluate complex instruction with eight features addressing the length, contents, and formats of input texts and task descriptions.

It is also empirically showed that complexity enhancement is necessary for performance improvement~\cite{zhao2023preliminary}. To increase the instruction complexity in SFT datasets, some works propose to incrementally augment existing instructions by adding nodes to semantic tree~\cite{zhao2023preliminary} or performing operations such as increasing reasoning, adding constraints, in-breadth evolving, deepening, and so on~\cite{xu2023wizardlm, luo2023wizardcoder, jiang2023followbench, sun2024conifer}.

\subsection{Data Quantity}
\label{subsec:stf-quantity}

Different with the acknowledged scaling laws of pretraining data, explorations of the relationship between scaling instruction data quantity and fine-tuned model performance diverge in two directions. In the earlier stage, researchers follow the observations in the pretraining of LLMs and argue that scaling up the instruction data quantity is crucial for success~\cite{wei2021finetuned, sanh2022multitask}. Recently, more works claim that data quality is more important than data quantity in the SFT of LLMs, and propose to scaling down the instruction datasets with limited high-quality data~\cite{zhou2023lima, chen2023maybe}. However, \citet{zhang2024scaling} propose a power-based multiplicative joint scaling law, showing that increased fine-tuning data could lead to improved model performance after achieving good results with limited data.

Addressing this conflict, several works attempt to analyze the scaling patterns for different tasks or different model abilities. A consensus of these works is that different abilities have different scaling patterns and develop at different paces. \citet{dong2023abilities} find that general ability can be enhanced with about 1,000 samples and improves slowly after then, while mathematical reasoning and code generation improve consistently with the increasing of instruction data amount. Similarly, \citet{yuan2023scaling} observe a log-linear relation between instruction data amount and models' mathematical reasoning performance, but stronger pretrained models improve less with more instruction data. Surprisingly, the empirical study of \citet{ji2023exploring} on 12 major real-world online user cases draws to an exactly opposite point. \citet{song2023dynamics} also show that some abilities have completely different patterns from others.

\subsection{Dynamic Data-Efficient Learning}
\label{subsec:sft-efficient}

While works discussed above focus more on the static management of instruction datasets without interaction with model fine-tuning, some works try to combine data selection with model fine-tuning, achieving data-efficient learning in a dynamic way.

\paragraph{Training affects data.} Some works propose to dynamically change the datasets along with the fine-tuning process. \citet{attendu-corbeil-2023-nlu} propose a dynamic data pruning method that periodically filters out unimportant examples during SFT using extended versions of EL2N metric~\cite{paul2021deep, fayyaz2022bert}. \citet{alshikh2023becoming} predict the responses as "answer-like or not" by a binary classifier, in order to measure LLMs' instruction-following ability and serve as an early-stopping criterion. \citet{Kung2023ActiveIT} conduct active task searching to select informative tasks based on prompt uncertainty and fine-tune in a loop.

\paragraph{Data affects training.} Instead of manipulating instruction datasets, some works propose special training strategies to accommodate the datasets. To mitigate forgetting and negative task impact, \citet{yin2023dynosaur} and \citet{wang2024inscl} treat task selection as a replay strategy in continual learning scenarios; \textit{DMT}~\cite{dong2023abilities} learns specialized and general abilities sequentially while keeping a small proportion of specialized data. To efficiently learn mixed-quality data acquired from LLMs with different level of abilities, \textit{OpenChat}~\cite{wang2023openchat} proposes \textit{C-RLFT} strategy that considers different data sources as coarse-grained reward labels; \citet{xu2023contrastive}, \citet{sun2024conifer} and \citet{kim2024strategic} propose to make the model progressively learn from easy to hard, respectively regarding different data quality, instruction complexity and task hardness.

\subsection{Relations Among Task composition, Data Quality and Data Quantity}
\label{sec:sft-relation}
Similar as in the pretraining stage, different aspects of supervised fine-tuning data management can affect model performance jointly. \citet{lu2023instag} analyze popular open-set SFT datasets using \textit{\#InsTag} and show that larger dataset sizes tend to be more diverse and induce higher performance. Current research on data selection tends to uniformly consider instruction quality and diversity~\cite{Bukharin2023DataDM, Xu2023RethinkingTI}. Since different model abilities have different scaling patterns as discussed in Section~\ref{subsec:stf-quantity}, more efficient task composition strategies are required to build stronger multi-task LLMs.

In summary, we provide a list of takeaways in Appendix~\ref{app:takeaways}. Some other aspects of data management are discussed in Appendix~\ref{app:others}.

\section{Challenges and Future Directions}
\label{sec:challenges}

The exploration of data management and its impact on LLM pretraining and SFT is still an ongoing task. In this section, we point out several challenges and corresponding future directions in the research of training data management for LLMs.

\paragraph{General data management framework} Although data management systems are proposed to compose various data recipes in either the pretraining or SFT stage of LLM~\cite{chen2023data, zhou2023oasis, sun2024integrated}, practitioners still need to spend efforts on organizing suitable datasets. A well-established general data management framework suitable for a broad range of applications is an urgent and worthy future direction in developing and promoting LLMs.

Beyond that, a more autonomous data management system is also needed to greatly save human efforts. To build such systems, LLMs might be leveraged and serve as different roles such as quality examinator, data augmentor, and so on.

\paragraph{Data debiasing and detoxifying} Current pretraining corpora and instruction datasets might contain harmful information and social biases, which lead to negative social impacts and undesirable model behavior. With the application of LLMs keeps extending to more demanding fields, the fairness and harmlessness of LLMs will become more and more innegligible. Hence, as one way to build ideal LLMs without biases and harmful output, debiasing and detoxifying of pretraining and instruction data is an important research direction.

\paragraph{Multimodal data management} Current research in data management mostly focuses on natural language processing. With the application of LLMs extending to modalities like vision, audio, etc., it is necessary to see the impacts of multimodal data management on the performance of fine-tuned multimodal LLMs.

\paragraph{Data management for LLM self-exploration} The ability to actively explore the unknown environment and tasks is one of the future perspectives in LLM development. Learning from large-scale interaction data requires efficient data management system to construct suitable datasets.

\paragraph{Efficient filtering for synthesized data} As data annotation requires intensive human labors and existing data will be exhausted, automatically synthesizing new data using LLMs is newly proposed as a promising solution~\cite{maini2024rephrasing, li2024synthetic}. In this process, efficient filtering for synthesized data is required to ensure its quality.

\paragraph{Fine-grained data ordering} Some works start to pay attention to the ordering of data in both the pretraining~\cite{gan2023ziya2, guo2024deepseek} and SFT stage~\cite{xu2023contrastive, yin2023dynosaur}. 
It is shown that more fine-grained data ordering could be beneficial to model performance improvement.

\paragraph{Conflicted data separation} In multi-task fine-tuning, negative impact of mixing data is observed and attribute to conflicts among different task data~\cite{dong2023abilities}. Thus, separating and effectively learning from conflicted data samples is a challenging problem in multi-task learning.

\section{Conclusions}

This paper overviews the training data management of LLMs. We discuss the \textit{pretraining} and \textit{supervised fine-tuning} stages of LLM successively and summarize the up-to-date research efforts according to the data management process of each stage. Finally, we highlight several challenges and future directions for LLM training data management. We hope this survey can provide insightful guidance for practitioners and inspire further research in efficient training data management for the development of LLMs.

\bibliography{anthology,custom}

\appendix

\section{Takeaways}
\label{app:takeaways}
In the pretraining stage of LLMs:
    \begin{itemize}
        \item The coverage of more domains and proper domain mixture ratio are important. Recently, researchers try to automatically find the proper domain mixture weights, which still show room for improvement.
        \item Large amount of data is widely considered critical, and proper data repetition may also bring positive impacts to model performance.
        \item Data quality control is necessary usually form an order, namely quality filtering, deduplication and toxicity filtering.. However, over-aggressive quality and toxicity filtering may lead to performance degradation and social biases, which is still under-explored. 
        \item Data diversity and temporal misalignment also have impacts on model performance, which call for future study.
    \end{itemize}
    
In the supervised fine-tuning stage of LLMs:
    \begin{itemize}
        \item Multitask fine-tuning is widely adopted nowadays. However, conflictions may exist among tasks and hinders the model abilities. Hence, dealing with negative task confliction is also calling for better answers. Ensembling multiple single-task experts instead of training one multitask model also arises as an new trend.
        \item Quality control are usually achieved through heuristics, human evaluation or LLMs as quality judges. Instruction diversity and complexity are also beneficial and enhanced by several works. The exploration of more diverse and complex instructions is still an open question.
        \item Works have shown that the SFT of LLM rely more on data quality than data quantity. However, digging deeper into the influence of data quantity, some researchers find that the learning of different tasks may require different amount of data.
        \item Instead of keep instruction datasets unchanged during fine-tuning, works propose to adjust the datasets dynamically through fine-tuning. Special fine-tuning strategies are also continually shown up to utilize the instruction data more efficiently.
    \end{itemize}

\section{Other Aspects of Data Management For LLMs}
\label{app:others}
\subsection{Social Bias}
\label{subsubsec:pt-bias}
Besides the marginalization of minority groups caused by data detoxifying mentioned in Section~\ref{subsubsec:toxic}, several works~\cite{kurita2019measuring, nangia2020crows, meade2021empirical, feng2023pretraining} find that pre-trained LLMs can capture social biases contained in the large amounts of training text. Evaluating on the C4.EN~\cite{raffel2020exploring} dataset, \citet{dodge2021documenting} recommend documenting the social biases and representational harms as well as excluded voices and identities in large web text corpora. Using a dataset of U.S. high school newspaper articles, \citet{gururangan-etal-2022-whose} also argue that the quality filters used for GPT-3~\cite{brown2020language} prefer newspapers published by larger schools located in wealthier, educated, and urban ZIP codes, leading to a language ideology. \citet{feng2023pretraining} conduct a comprehensive case study focusing on the effects of media political biases in the pretraining corpus on the fairness of hate speech detection and misinformation detection w.r.t. partisan leanings and how it is propagated to language models even further to downstream tasks.

As addressed in previous research, there is still a large gap between current prominent LLMs and ideal LLMs without social biases. Many questions are worth exploring, such as how to mitigate the potential biases in pretraining datasets, the existence of bias in the SFT datasets, and whether it is feasible to reduce social bias through SFT.

\subsection{Prompt Design}
Current instructions are either heuristically designed by human~\cite{wang2022super, kopf2023openassistant} or synthetically generated by prominent models~\cite{peng2023instruction, ding2023enhancing}. 
The choice of prompts might cause significant model performance variation~\cite{gonen2022demystifying, weber2023mind}. 
Early attempts include manual reformulation of prompts into the ones easier to follow for language models~\cite{mishra-etal-2022-reframing}, and choosing prompts with the lowest perplexity to get the most significant gains in model performance~\cite{gonen2022demystifying}. Recently, \citet{liang2023exploring} develop a format transfer framework \textit{UIT} to transfer instructions from different datasets into unified formats automatically. 

Some works focus on studying the impact of prompt phrasing. \citet{khashabi-etal-2022-prompt} surprisingly find that the discretized interpretation of continuous prompts is not always consistent with the discrete prompts describing the same task as heuristically expected. 
\citet{yin-etal-2023-read} find that removing the descriptions of task output, especially the label information, might be the only reason for performance degradation. They also propose an automatic task definition compression algorithm to remove almost half or more of the tokens while improving model performance. \citet{kung-peng-2023-models} also remove all semantic components in task definitions but the output space information. They achieve comparable model performance using the modified task definitions and delusive examples containing incorrect input-output mappings. Based on their experiment results, they cast doubts on the performance gain of fine-tuned models and state that the model may only learn superficial patterns during instruction tuning. 

Besides the choice of phrasing, the generation source of prompts is another factor in prompt design. \citet{gudibande2023false} raise questions on fine-tuning a weaker language model on outputs of a stronger model and find that the imitation model might adapt to mimic the stronger model's style but not its functionality.
Similarly, \citet{song2023dynamics} also observe that human-designed data can outperform synthetically generated data from GPT-4~\cite{openai2023gpt4} to a relatively large extent.

\subsection{Hallucinations}
Despite their strong power, LLMs are notorious for their hallucinations, i.e. the generation of input-, context- or fact-conflicting contents~\cite{zhang2023hallucination}. Several works in hallucination trace down the occurrence of hallucination to the lack of pertinent knowledge and the internalization of false knowledge from the pretraining corpora~\cite{li2022pre, mckenna2023sources, dziri-etal-2022-origin}. To mitigate hallucination, the curation of pretraining corpora is adopted by many LLMs, mainly focusing on the extracting of high-quality data, e.g., GPT-3~\cite{brown2020language}, Llama 2~\cite{touvron2023llama2}, and Falcon~\cite{penedo2023refinedweb}. The manually curated~\cite{zhou2023lima} and automatically selected~\cite{chen2023alpagasus, cao2023instruction, lee2023platypus} high-quality instruction data are also experimentally shown to be effective in reducing hallucination during the SFT stage. It can be seen from the previous research that data management in both the pretraining and SFT stages can be a promising solution to hallucination.

\section{Related Surveys}
As LLMs draw more and more attention, a handful of surveys have been published or preprinted addressing different aspects of their development. Related to our work, several of them also include parts of the data preparation process in the pretraining or SFT of LLM. \citet{zhao2023survey} review the development of LLMs and the latest advancements covering a wide range of topics. \citet{yang2023harnessing} also provide an overview of the LLM evolution and discuss the related techniques from model, data, and downstream tasks. Also concentrating on data, \citet{zha2023data} introduce data-centric AI and its related tasks and methods for general machine learning models instead of LLMs. \citet{zhang2023instruction} survey the instruction tuning of LLMs and its related methodologies, data construction, applications, and so on. \citet{wang2023aligning} review the technologies aligning LLMs with human expectations including data collection, training methodologies, and model evaluation.

Unlike previous surveys, this survey provides a systematic and detailed overview of data management at both the pretraining and SFT stages of LLMs. We focus on the proper organization of training datasets and discuss recent research addressing the effects of different data management strategies, the evaluation of curated training datasets, and the latest advances in training data management strategies, providing a guiding resource for practitioners aiming to build powerful LLMs through efficient data management.

\section{Comparison of Data Management Strategies Used by Representative LLMs}
\label{app:comparison}
We provide two summary tables, Table~\ref{tab:pretrain} for pretrained LLMs and Table~\ref{tab:sft} for SFT LLMs, with better and clearer comparison of the data management strategies used by current representative LLMs.

\begin{table*}[t!]
    \centering
    \small
    \begin{tabular}{c|clllll}
         Pretrained LLMs & \makecell{Open-\\sourced} & Quantity & Deduplication & \makecell{Quality\\Filters}& \makecell{Toxicity\\Filters} & Domian Composition \\
        \hline
         \makecell{T5\\~\cite{raffel2020exploring}}& $\surd$& 750GB& N-gram& Heuristic& Heuristic& 99\% Web, < 1\% Wiki \\[10pt]
         \makecell{GPT-3\\~\cite{brown2020language}}&& \makecell[l]{499B\\tokens} & \makecell[l]{MinHash,\\LSH}& Classifier&& \makecell[l]{82\% Web, 16\% Books,\\3\% Wiki} \\[10pt]
         \makecell{GLaM\\~\cite{du2022glam}}&& \makecell[l]{1.6T\\tokens} && Classifier && \makecell[l]{46\% Web, 28\% Dialog,\\20\% Books, 6\% Wiki} \\[10pt]
         \makecell{LaMDA\\~\cite{thoppilan2022lamda}}&& \makecell[l]{1.56T\\words} &&& & \makecell[l]{50\% Dialog, 25\% Web,\\12.5\% Wiki,\\12.5\% Code} \\[10pt]
         \makecell{Chinchilla\\~\cite{hoffmann2022empirical}} && \makecell[l]{1.4T\\tokens} & \makecell[l]{N-gram,\\Doc-level}& Heuristic& Heuristic& \makecell[l]{65\% Web, 30\% Books,\\4\% Code, 1\% Wiki} \\[10pt]
         \makecell{AlphaCode\\~\cite{li2022competition}} && 715.1GB& Doc-level& Heuristic&& 100\% Code \\[5pt]
         \makecell{GLM\\~\cite{zeng2022glm}}& $\surd$& \makecell[l]{400B\\tokens} &&&& \makecell[l]{50\% Pile,\\50\% Chinese Web data} \\[10pt]
         \makecell{BLOOM\\~\cite{le2023bloom}}& $\surd$& \makecell[l]{1.61TB\\text} & \makecell[l]{SimHash,\\Substring\\clustering} & Heuristic& Heuristic& \makecell[l]{60\% Web, 10\% Books,\\10\% Code,\\10\% Academic,\\5\% Dialog, 5\% Wiki} \\[20pt]
         \makecell{PaLM\\~\cite{anil2023palm}}&& \makecell[l]{780B\\tokens} & \makecell[l]{Levenshtein\\distance}& \makecell[l]{Heuristic,\\Classifier} & Classifier& \makecell[l]{50\% Dialog, 28\% Web,\\13\% Books, 5\% Code,\\4\% Wiki} \\[20pt]
         \makecell{LLaMA\\~\cite{touvron2023llama}}& $\surd$& \makecell[l]{1.4T\\tokens} & \makecell[l]{Line-level,\\Book-level}& \makecell[l]{Heuristic,\\Classifier} & Classifier& \makecell[l]{82\% Web, 4.5\% Code,\\4.5\% Wiki,\\4.5\% Books,\\2.5\% Academic,\\2\% Dialog} \\[20pt]
         \makecell{Mistral\\~\cite{jiang2023mistral}}& $\surd$& - & - & - & - & - \\[10pt]
         \makecell{phi-1/1.5\\~\cite{gunasekar2023textbooks}\\~\cite{li2023textbooks}} & $\surd$& \makecell[l]{7B\\tokens}  && Classifier&& \makecell[l]{99\% Academic,\\<1\% Code} \\[20pt]
         \makecell{phi-2\\~\cite{2023phi2}} & $\surd$ & \makecell[l]{1.4B\\tokens} && Classifier&& \\[10pt]
         \makecell{GPT-4\\~\cite{openai2023gpt4}} && - & - & - & - & -  \\[10pt]
         \makecell{LLaMA 2\\~\cite{touvron2023llama2}}& $\surd$ & \makecell[l]{2.0T\\tokens} &&& Heuristic & \\[10pt]
         \makecell{QWen\\~\cite{bai2023qwen}} & $\surd$& \makecell[l]{3T\\tokens} & \makecell[l]{Exact Match,\\MinHash,\\LHS} & \makecell[l]{Heuristic,\\Classifier} & Classifier & \makecell[l]{Web, Books,\\Codes, Academic} \\[5pt]
         \makecell{Deepseek LLM\\~\cite{bi2024deepseek}} & $\surd$ & - & - & - & - & - \\
    \end{tabular}
    \caption{The data management strategies used by representative pretrained models. The blank units mean no specific design of corresponding strategies according to the original papers. The '-' means the data management process is not released. Part of the data is adopted from~\cite{longpre2023pretrainer}}
    \label{tab:pretrain}
\end{table*}

\begin{table*}[t!]
    \centering
    \small
    \begin{tabular}{c|lllllll}
        SFT LLMs & Dataset& Quantity & \makecell[l]{Quality\\Control} & \makecell[l]{Diversity\\Control}  & \makecell{Complexity\\Enhancing} & \makecell{No. of\\Tasks} & \makecell[l]{Task\\Balancing} \\
        \hline
        \makecell{Tk-Instruct\\~\cite{wang2022super}} & NIv2 & 5M & \makecell[l]{Heuristic\\Human}&&& 1616 & \makecell[l]{Limited\\instances\\per task} \\[20pt]
        \makecell{Flan-T5\\~\cite{longpre2023flan}} & Flan 2022 & 15M && \makecell[l]{Input\\Inversion} && 1836 & \makecell[l]{Experiments\\intuitions} \\[10pt]
        \makecell{OPT-IML\\~\cite{iyer2022opt}} & \makecell[l]{OPT-IML\\Bench} & 18M &&&& 2000& Experiments \\[10pt]
        \makecell{Alpaca\\~\cite{alpaca}} & Alpaca & 52K & Heuristic & \makecell[l]{ROUGE-L\\similarity} && 80& \\[10pt]
        \makecell{Vicuna\\~\cite{vicuna2023}} & ShareGPT & 70K & Heuristic &&& \\[10pt]
        \makecell{LIMA\\~\cite{zhou2023lima}} & LIMA & 1K& \makecell[l]{Heuristic\\Human}& \makecell[l]{Heuristic,\\Human}&&& \\[10pt]
        \makecell{Dolly\\~\cite{DatabricksBlog2023DollyV2}} & dolly-15k & 15K & Human &&&& \\[10pt]
        \makecell{Orca\\~\cite{mukherjee2023orca}} & \makecell[l]{sampled\\Flan 2022} & 5M &&& \makecell[l]{Chat-GPT/\\GPT-4\\augmentation} && \\[20pt]
        \makecell{WizardLM\\~\cite{xu2023wizardlm}\\WizardCoder\\~\cite{luo2023wizardcoder}\\WizardMath\\~\cite{luo2023wizardmath}} & \makecell[l]{WizardLM\\WizardCoder\\WizardMath} & 250K && Evol-Instruct & Evol-Instruct && \\[30pt]
        \makecell{AlpaGasus\\~\cite{chen2023alpagasus}}  & AlpaGasus & 9K & \makecell[l]{Chat-GPT\\grading} &&&& \\[10pt]
        \makecell{Platypus\\~\cite{lee2023platypus}}& \makecell[l]{Open-\\Platypus}& 25K & \makecell[l]{Dedup,\\Heuristic} &&&& \\[10pt]
        \makecell{OpenChat\\~\cite{wang2023openchat}} & ShareGPT& 6K & C-RLFT &&&& \\[10pt]
        \makecell{MAmmoTH\\~\cite{yue2023mammoth}} & MathInstruct & 260K &&&& \makecell[l]{7 math\\fields} & \makecell[l]{Combining\\CoT and PoT} \\[10pt]
    \end{tabular}
    \caption{The data management strategies used by representative supervised finetuned models. The blank units mean no specific design of corresponding strategies according to the original papers. "NIv2" is the abbreviation for "Super-NaturalInstructions". "Dedup" is the abbreviation for "Deduplication".}
    \label{tab:sft}
\end{table*}

\section{Taxonomy}
\label{sec:app-tax}

The full taxonomy of research discussed in this survey is illustrated in Figure~\ref{fig:framework}

\tikzstyle{my-box}=[
    rectangle,
    draw=hidden-draw,
    rounded corners,
    text opacity=1,
    minimum height=1.5em,
    minimum width=5em,
    inner sep=2pt,
    align=center,
    fill opacity=.5,
    line width=0.8pt,
]
\tikzstyle{leaf}=[my-box, minimum height=1.5em,
    fill=hidden-pink!80, text=black, align=left,font=\normalsize,
    inner xsep=2pt,
    inner ysep=4pt,
    line width=0.8pt,
]
\begin{figure*}[t!]
    \centering
    \resizebox{\textwidth}{!}{
        \begin{forest}
            forked edges,
            for tree={
                grow=east,
                reversed=true,
                anchor=base west,
                parent anchor=east,
                child anchor=west,
                base=center,
                font=\large,
                rectangle,
                draw=hidden-draw,
                rounded corners,
                align=left,
                text centered,
                minimum width=4em,
                edge+={darkgray, line width=1pt},
                s sep=3pt,
                inner xsep=2pt,
                inner ysep=3pt,
                line width=0.8pt,
                ver/.style={rotate=90, child anchor=north, parent anchor=south, anchor=center},
            },
            where level=1{text width=13em,font=\normalsize,}{},
            where level=2{text width=14em,font=\normalsize,}{},
            where level=3{text width=11em,font=\normalsize,}{},
            [
                Data Management, ver
                [
                    Pretraining (\S \ref{sec:pretraining}), ver
                    [
                        Domain Composition (\S \ref{subsec:pt-composition})
                        [
                            \citet{longpre2023pretrainer}{, }\citet{nijkamp2023codegen2}{, }\citet{shen2023slimpajama}{, } \\
                            \citet{xie2023data}{, }\citet{xie2023doremi}{, }\citet{fan2023doge}{, }\citet{ye2024data}{, }\\
                            \cite{xia2023sheared}, leaf, text width=33em
                        ]
                    ]
                    [
                        Data Quantity (\S \ref{subsec:pt-quantity})
                        [
                            Scaling Laws
                            [
                                \citet{kaplan2020scaling}{, }\citet{hoffmann2022empirical}{, }\citet{su2024unraveling}, leaf, text width=33em
                            ]
                        ]
                        [
                            Data Repetition
                            [
                                \citet{villalobos2022will}{, }\citet{muennighoff2023scaling}{, }\citet{hernandez2022scaling}{, }\\
                                \citet{xue2023repeat}{, }\citet{tirumala2023d4}, leaf, text width=33em
                            ]
                        ]
                    ]
                    [
                        Data Quality (\S \ref{subsec:pt-quality})
                        [
                            Quality  Filtering
                            [
                                \citet{gao2021empirical}{, }\citet{kreutzer-etal-2022-quality}{, }\citet{gunasekar2023textbooks}{, } \citet{li2023textbooks}{, } \\
                                \citet{penedo2023refinedweb}{, }\citet{marion2023less}{, }\citet{longpre2023pretrainer}{, } 
                                \\
                                \citet{kaddour2023minipile}{, }\citet{2023phi2}{, }\citet{gan2023ziya2}{, }\\
                                \citet{wettig2024qurating}, leaf, text width=33em
                            ]
                        ]
                        [
                            Deduplication
                            [
                                \citet{lee2021deduplicating}{, }\citet{kandpal2022deduplicating}{, }\citet{silcock2022noise}{, } \\
                                \citet{abbas2023semdedup}, leaf, text width=33em
                            ]
                        ]
                        [
                            Toxicity Filtering
                            [
                                \citet{luccioni2021s}{, }\citet{xu2021detoxifying}{, }\citet{welbl2021challenges}{, } \\
                                \citet{longpre2023pretrainer}, leaf, text width=33em
                            ]
                        ]
                        [
                            Diversity \& Age
                            [
                                \citet{lee2023beyond}{, }\citet{maharana2023d2}{, }\citet{longpre2023pretrainer}, leaf, text width=33em
                            ]
                        ]
                        [
                            Social Bias*
                            [
                                \citet{dodge2021documenting}{, }\citet{meade2021empirical}{, }\citet{gururangan-etal-2022-whose} \\
                                \citet{feng2023pretraining}, leaf, text width=33em
                            ]
                        ]
                        [
                            Hallucinations*
                            [
                                \citet{li2022pre}{, }\citet{mckenna2023sources}{, }\citet{dziri-etal-2022-origin}, leaf, text width=33em
                            ]
                        ]
                    ]
                    [
                        Relations Among \\Different Aspects (\S \ref{sec:pt-relation})
                        [
                            \citet{ge2024data}{, }\citet{goyal2024scaling}{, }\citet{bi2024deepseek}{, }\citet{shen2023slimpajama}{, }\\
                            \citet{longpre2023pretrainer}, leaf, text width=33em
                        ]
                    ]
                ]
                [
                    Supervised Fine-Tuning (\S \ref{sec:sft}), ver
                    [
                        Task Composition (\S \ref{subsec:sft-composition})
                        [
                            \citet{wei2021finetuned}{, }\citet{wang2022super}{, }\citet{sanh2022multitask}{, }\citet{chung2022scaling}{, } \\
                            \citet{longpre2023flan}{, }\citet{jang2023exploring}{, }\citet{chen2023maybe}{, }\citet{xia2024less} \\ 
                            \citet{dong2023abilities}{, }\citet{iyer2022opt}{, }\citet{wang2023far}{, }\citet{ivison-etal-2023-data}{, }\\
                            \citet{lee2024instruction}, leaf, text width=33em
                        ]
                    ]
                    [
                        Data Quality (\S \ref{subsec:sft-quality})
                        [
                            Instruction Quality
                            [
                                \citet{chia2023instructeval}{, }\citet{zhou2023lima}{, }\citet{Li2023FromQT}{, }\citet{Li2024SelectiveRS} \\
                                \citet{ding2023enhancing}{, }\citet{wang2023harnessing}{, }\citet{li2023self}{, }\citet{zhou2023lobass}{, }\\
                                \citet{cao2023instruction}{, }\citet{madaan2023self}{, }\citet{Du2023MoDSMD}{, }\citet{Li2024SuperfilteringWD} \\
                                \citet{lu2023self}{, }\citet{selfee2023}{, }\citet{chen2023alpagasus}{, }\citet{Li2023TunaIT}{, } \\
                                \citet{Li2023OneSL}{, }\citet{Bhatt2024AnED}{, }\citet{chen2024automated}{, }\\
                                \citet{yang2024smalltolarge}{, }
                                \citet{mekala2024smaller}{, }\citet{he2024shed}{, }\citet{liu2024selectit}, leaf, text width=33em
                            ]
                        ]
                        [
                            Instruction Diversity
                            [
                                \citet{ding2023enhancing}{, }\citet{zhou2023lima}{, }\citet{Bukharin2023DataDM} \\
                                \citet{alpaca}{, }\citet{lu2023instag}{, }\citet{wang-etal-2023-self-instruct}{, } \\
                                \citet{wan2023explore}{, }\citet{wu2023self}{, }\citet{ge2024clustering}{, }\citet{huang2024multi}, leaf, text width=33em
                            ]
                        ]
                        [
                            Instruction Complexity
                            [
                                \citet{lu2023instag}{, }\citet{xu2023wizardlm}{, }\citet{luo2023wizardcoder}{, }\citet{mukherjee2023orca}{, } \\
                                \citet{zhao2023preliminary}{, }\citet{he2023can}{, }\citet{jiang2023followbench}{, }\citet{sun2024conifer}\\
                                \citet{he2024complex}, leaf, text width=33em
                            ]
                        ]
                        [
                            Prompt Design*
                            [
                                \citet{mishra-etal-2022-reframing}{, }\citet{khashabi-etal-2022-prompt}{, }\citet{gonen2022demystifying}{, } \\
                                \citet{yin-etal-2023-read}{, }\citet{kung-peng-2023-models}{, }\citet{liang2023exploring}{, } \\
                                \citet{weber2023mind}{, }\citet{gudibande2023false}{, }\citet{song2023dynamics}, leaf, text width=33em
                            ]
                        ]
                        [
                            Hallucinations*
                            [
                                \citet{zhou2023lima}{, }\citet{chen2023alpagasus}{, }\cite{cao2023instruction}{, }\citet{lee2023platypus}, leaf, text width=33em
                            ]
                        ]
                    ]
                    [
                        Data Quantity (\S \ref{subsec:stf-quantity})
                        [
                            \citet{ji2023exploring}{, }\citet{zhou2023lima}{, }\citet{yuan2023scaling}{, }\citet{zhang2024scaling} \\
                            \citet{chen2023maybe}{, }\citet{dong2023abilities}{, }\citet{song2023dynamics}, leaf, text width=33em
                        ]
                    ]
                    [
                        Dynamic \\Data-Efficient Learning (\S \ref{subsec:sft-efficient})
                        [
                            Training Affects Data
                            [
                                \citet{attendu-corbeil-2023-nlu}{, }\citet{alshikh2023becoming}{, }\citet{Kung2023ActiveIT}, leaf, text width=33em
                            ]
                        ]
                        [
                            Data Affects Training
                            [
                                \citet{yin2023dynosaur}{, }\citet{wang2023openchat}{, }\citet{dong2023abilities}{, }\\
                                \citet{xu2023contrastive}{, }\citet{wang2024inscl}{, }\citet{sun2024conifer}{, }\citet{kim2024strategic}, leaf, text width=33em
                            ]
                        ]
                    ]
                    [
                        Relations Among \\Different Aspects (\S \ref{sec:pt-relation})
                        [
                            \citet{lu2023instag}{, }\citet{Bukharin2023DataDM}{, }\citet{Xu2023RethinkingTI}, leaf, text width=33em
                        ]
                    ]
                ]
            ]
        \end{forest}
    }
    \caption{Taxonomy of research in data management for pretraining and supervised fine-tuning of Large Language Models (LLM).}
    \label{fig:framework}
\end{figure*}
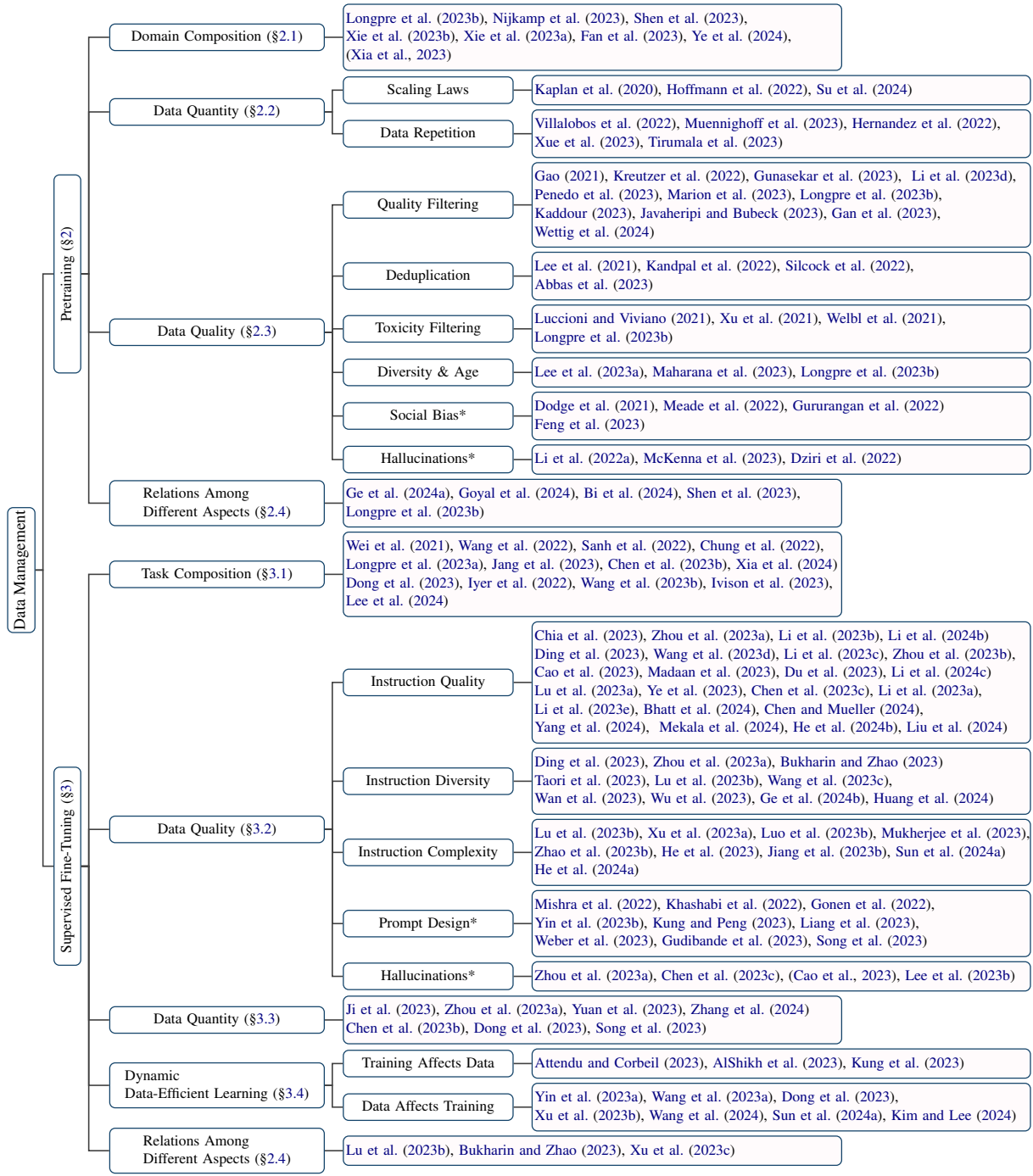

\end{document}